\documentclass[10pt,twocolumn,letterpaper]{article}

\usepackage{cvpr}
\usepackage{times}
\usepackage{epsfig}
\usepackage{graphicx}
\usepackage{amsmath}
\usepackage{dsfont}
\usepackage{amssymb}
\usepackage{placeins}
\usepackage[hidelinks]{hyperref}

\newcommand{\fig}[1]{Figure~\ref{fig:#1}}
\newcommand{\sect}[1]{Section~\ref{sect:#1}}

\newcommand{\cvprPaperID}{5214}

\begin{document}
\cvprfinalcopy
% \pagenumbering{gobble}
\title{StylePeople: A Generative Model of Fullbody Human Avatars}

% The \author macro works with any number of authors. There are two
% commands used to separate the names and addresses of multiple
% authors: \And and \AND.
%
% Using \And between authors leaves it to LaTeX to determine where to
% break the lines. Using \AND forces a line break at that point. So,
% if LaTeX puts 3 of 4 authors names on the first line, and the last
% on the second line, try using \AND instead of \And before the third
% author name.

% \author{
%   David S.~Hippocampus\thanks{Use footnote for providing further
%     information about author (webpage, alternative
%     address)---\emph{not} for acknowledging funding agencies.} \\
%   Department of Computer Science\\
%   Cranberry-Lemon University\\
%   Pittsburgh, PA 15213 \\
%   \texttt{hippo@cs.cranberry-lemon.edu}  \\
%   % examples of more authors
%   % \And
%   % Coauthor \\
%   % Affiliation \\
%   % Address \\
%   % \texttt{email} \\
%   % \AND
%   % Coauthor \\
%   % Affiliation \\
%   % Address \\
%   % \texttt{email} \\
%   % \And
%   % Coauthor \\
%   % Affiliation \\
%   % Address \\
%   % \texttt{email} \\
%   % \And
%   % Coauthor \\
%   % Affiliation \\
%   % Address \\
%   % \texttt{email} \\
% }

\author{
Artur Grigorev $^{1,2}$\thanks{equal contribution}   , \\
\texttt{a.grigorev@samsung.com}  
\and
Karim Iskakov $^{1}$\footnotemark[1]  , \\
\texttt{kar.iskakov@gmail.com} 
\and
Anastasia Ianina $^1$, \\
\texttt{a.ianina@samsung.com}  
\and
Renat Bashirov $^1$, \\
\texttt{r.bashirov@samsung.com} 
\and
Ilya Zakharkin $^{1,2}$, \\
\texttt{ilyazaharkin@gmail.com} 
\and
Alexander Vakhitov $^1$, \\
\texttt{alexander.vakhitov@gmail.com} 
\and
Victor Lempitsky $^{1,2}$ \\
\texttt{v.lempitsky@samsung.com}
\\
\\ $^1$ Samsung AI Center, Moscow 
\\ $^2$ Skolkovo Institute of Science and Technology, Moscow
\vspace{-20pt}
}

\maketitle

\begin{abstract}
We propose a new type of full-body human avatars, which combines parametric mesh-based body model with a neural texture. We show that with the help of neural textures, such avatars can successfully model clothing and hair, which usually poses a problem for mesh-based approaches. We also show how these avatars can be created from multiple frames of a video using backpropagation. We then propose a generative model for such avatars that can be trained from datasets of images and videos of people. The generative model allows us to sample random avatars as well as to create dressed avatars of people from one or few images. The code for the project is available at \href{https://saic-violet.github.io/style-people}{\texttt{saic-violet.github.io/style-people}}.
\end{abstract}

\section{Introduction}
\label{sect:intro}

Creating realistically-looking and articulated human avatars is a challenging task with many applications in telepresence, gaming, augmented and virtual reality. In recent years, sophisticated and powerful models of ``naked'' people that model shape of the body including facial and hand deformations have been developed~\cite{Pavlakos19,Xiang19}. These models are based on mesh geometry and are learned from several existing datasets of body scans.
Clothing and hair literally add an extra layer of complexity on top of body modeling, they are even more challenging for appearance modeling, and accurate 3D data for these elements are scarce and hard to obtain. And yet creating realistic avatars is not possible without modeling of these elements. 

Here, we propose a new approach that we call \textit{neural dressing} that allows to create 3D realistic full body avatars from videos and in a few-shot mode (from one or several images). Similarly to previous works, this approach uses deformable meshes (specifically, SMPL-X model~\cite{Pavlakos19}) to model and animate body geometry in 3D. On top of the body mesh, the approach superimposes a multi-channel neural texture~\cite{thies2019deferred} that is processed by a rendering network in order to generate images of a full-body avatar with clothing and hair. Our first contribution is thus to show that the combination of deformable mesh models and neural textures (\textit{neural dressing}) can model appearance of full-body avatars with loose clothing and hair well and to account for the geometry missing in parametric body models. 

\begin{figure*}
    \centering
    \includegraphics[width=0.9\textwidth]{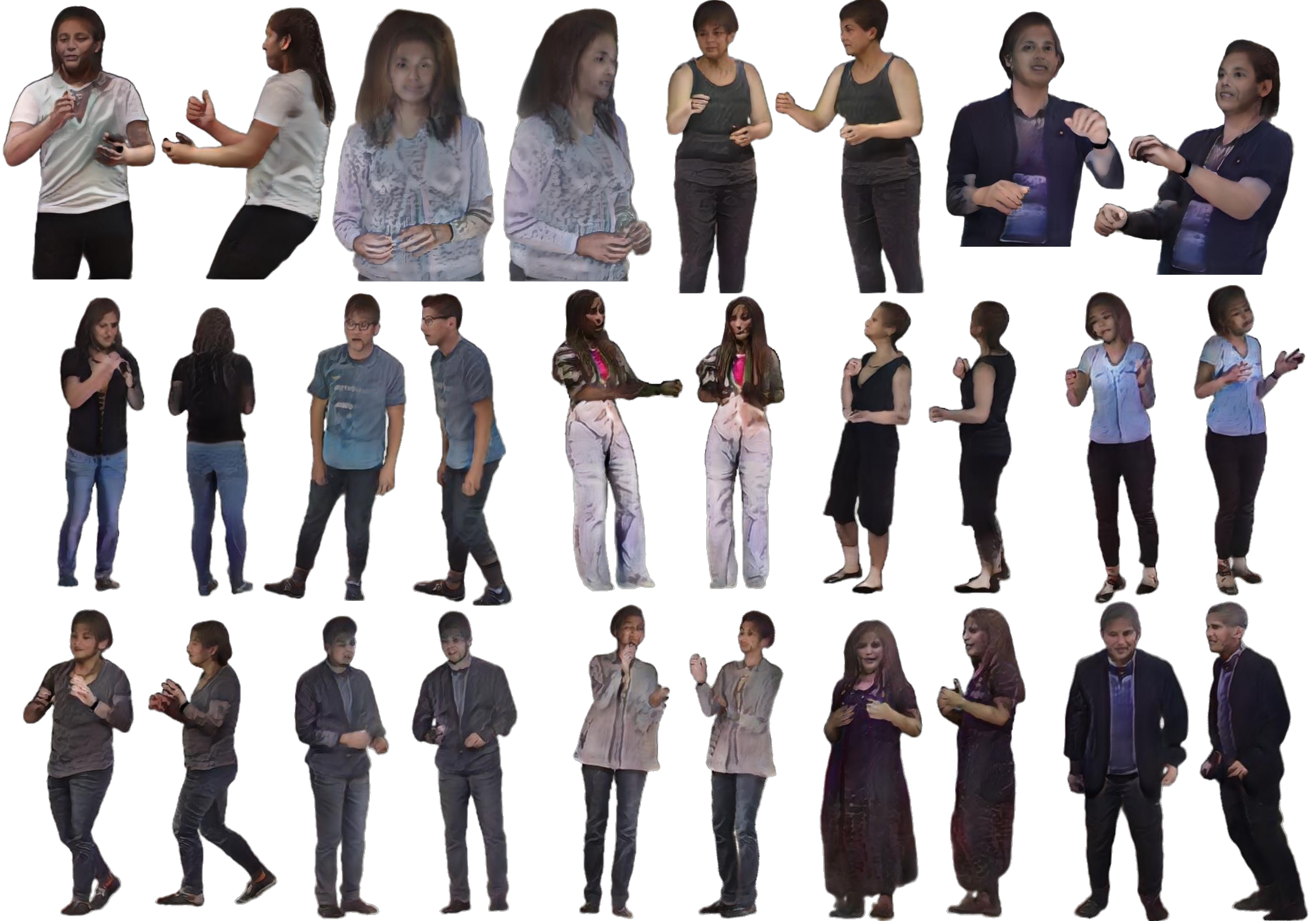}
    \caption{\textbf{Style people}, i.e.~random samples from our generative models of human avatars (truncation factor 0.8) . Each avatar is shown from two different viewpoints. The samples show diversity in terms of clothing and demographics. Loose clothing and hair are present. }
    \label{fig:teaser}
\end{figure*}

As our second contribution, we build a generative model of full-body avatars. The key component of the new model is a generative network for neural body texture. The generative network is derived from StyleGANv2~\cite{karras2019stylegan2} generator.
% with significant modifications. 
To build the complete model, we thus incorporate neural texture synthesis, mesh rendering, and neural rendering into the joint generation process, which is trained in an adversarial fashion on a large-scale dataset of full-body images. We also address the need to ensure that avatars have consistent appearance across variety of poses and camera positions. This is ensured by adding an additional discriminator network and by modifying the training process accordingly. 

Using the resulting generative model, we can sample new realistic 3D ``artificial humans'' (\textit{StylePeople} -- \fig{teaser}). Furthermore, the availability of a generative model allows us to build avatars for existing people by fitting the model to a single image or few images of a given person. As the generative model is heavily over-parameterized, we investigate how the fitting process can be regularized. In the experiments, we then compare our model to previous approaches to few-shot avatar modeling.

\section{Related work} 

An early generative model for people in clothing~\cite{lassner2017generative} translates renderings of a parametric body model to human parsing (semantic segmentation) maps, which are subsequently translated into human images. The pose-guided human image generation~\cite{ma2017pose} methods synthesize human images in new poses or with new clothes from a single image by either relying on feature warping in pose-to-image translation~\cite{siarohin2018deformable,esser2018variational,dong2018soft,han2019clothflow} or on predicting surface coordinates in the source and target frames ~\cite{alp2018densepose} and sampling RGB texture maps~\cite{neverova2018dense,grigorev2018coordinate}.  In~\cite{mir2020learning} the textures are transfered from clothing images to 3D garments worn on top of SMPL~\cite{loper2015smpl} using only shape information and ignoring texture. 

In a video-to-video setting, \cite{wang2018video} trains a system that transforms a sequence of body poses into corresponding temporally-consistent monocular videos. Likewise, \cite{shysheya2019textured} proposes to translate renderings of body joints to clothed body surface coordinates and use the latter to sample the learnable RGB texture stack achieving better visual quality. The geometry-texture decomposition helps occlusion reasoning and generalization to unseen poses, however surface coordinate regression mistakes result in visual artifacts, especially for the face and hands. Our approach goes one step further by switching from body coordinate regression to fitting a deformable  model to images, improving alignment and geometric consistency. 

The works~\cite{alldieck2018video,alldieck2019learning,alldieck2019tex2shape,bhatnagar2019multi,lazova2019360,huang2020arch} take a single or multiple images of a person and produce a texture and displacement map for a body model, which later can be rendered from any viewpoint and in arbitrary pose. However the quality of rendering the human avatar in this case is limited by the classic graphical pipeline, while in our case a rendering network allows to increase photorealism.

Recently, approaches that rely on implicit functions to create 3D model of human bodies~\cite{saito2019pifu,saito2020pifuhd} emerged. They allow to generate detailed shapes of clothed human body. Although, they are restricted to manipulate those shapes only in terms of viewpoint since they lack structural information and can not render fitted humans in new poses.

\begin{figure*}
    \centering
    \includegraphics[width=0.9\textwidth]{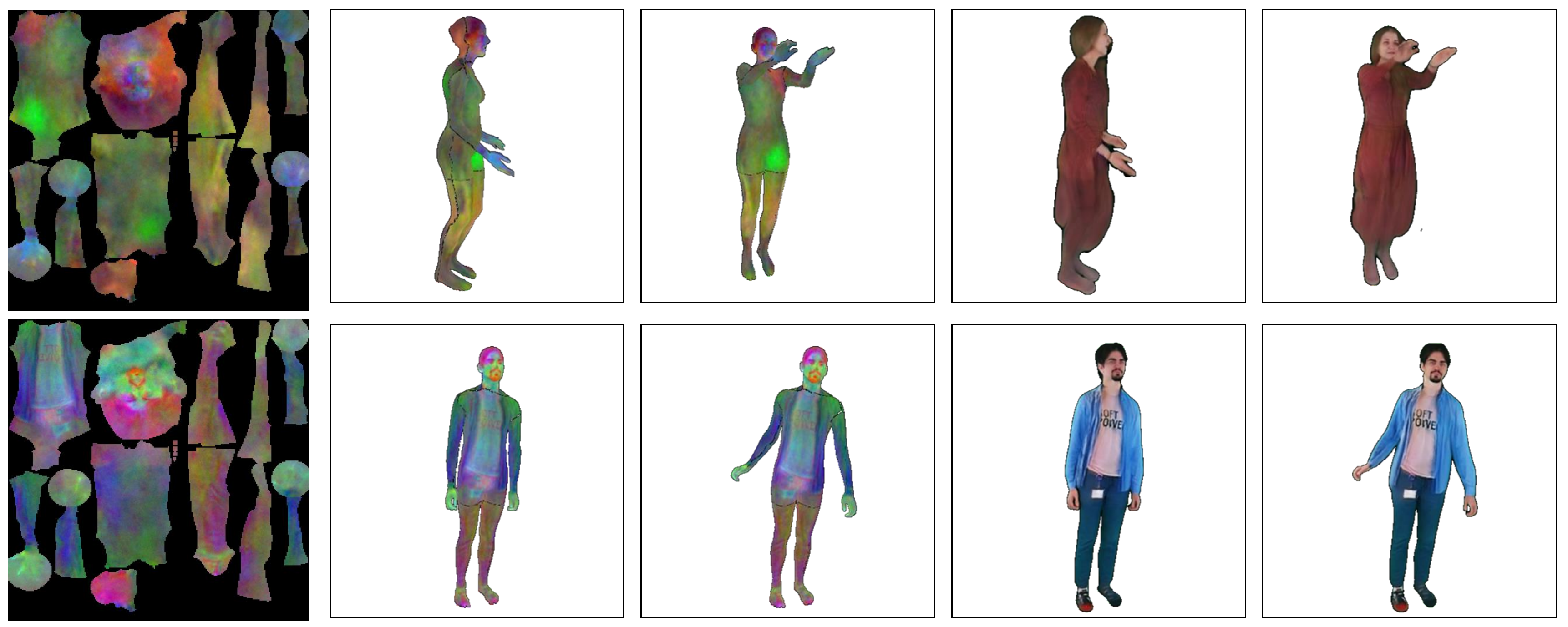}
    \caption{The neural dressing model. The texture of an avatar (left column; first three PCA components are visualized) is superimposed on the body mesh modeled with SMPL-X~\cite{Pavlakos19} (columns 2,3). Processing the rasterizations with the rendering network generates the renders (columns 4,5). Note that the model can handle loose clothing and long hair as well as complex garment textures gracefully.}
    \label{fig:dressing}
\end{figure*}

We build our approach on a powerful {\it deferred neural rendering} model of~\cite{thies2019deferred} proposing learnable neural textures for meshes, which encode visual appearance and can be rendered using a standard graphics pipeline and then translated into photorealistic images of objects and scenes. In our system, the neural textures also implicitly encode non-modelled geometry, e.g.\ clothing and hair. While the system of Thies~et~al.\ needs to be trained from scratch requiring at least several dozens of images per scene, we propose a generative model of neural textures for people rendering, trained on a new large dataset of annotated in-the-wild images, which allows few-shot synthesis of a photorealistic human avatar. The most recent approaches \cite{wang2019few, zablotskaia2019dwnet, liu2019liquid,siarohin2019first} report pose-to-image video-to-video translation in the few-shot regime, when the target appearance is specified by several images. Our approach leads to superior visual quality in the same setting by relying on a better geometric proxy as well as on the geometry-texture decomposition.

Finally, our StylePeople model builds on top of the recent years of advances in modeling high-resolution 2D images (\cite{Karras18,Brock18,Karras19}), in particular StyleGANv2 model~\cite{karras2019stylegan2}. Tewari~et~al.~\cite{tewari2020stylerig} propose to use StyleGAN model for rig-like control over face features. However, they can not control parts of the scene that are not explained by the parametric face model (e.g.\ hair style). The pairwise discriminator that ensures consistency between poses of the same avatar in our work has been recently proposed in a similar extension for StyleGAN model in \cite{Logacheva20} for the task of video synthesis (while in general, the use of multi-frame discriminators for consistent video synthesis can be traced to~\cite{Tulyakov18}). The inference ideas we apply to invert our generative model in few-shot learning mode are based on extended generator space~\cite{Karras19,Abdal19}, learning the auxiliary encoders~\cite{karras2019stylegan2,Logacheva20}, using discriminator feature matching loss~\cite{Pan20}, as well as fine-tuning generators to achieve the best fit~\cite{Zakharov19,Bau19,Pan20}.

\newcommand{\p}{{\mathbf p}}
\newcommand{\s}{{\mathbf s}}
\newcommand{\N}{{\mathbf N}}
\newcommand{\w}{{\mathbf w}}
\newcommand{\z}{{\mathbf z}}
\newcommand{\R}{\mathds{R}}
\newcommand{\Gen}{\mathcal{G}}
\newcommand{\Enc}{\mathcal{E}}
\newcommand{\Ren}{\mathcal{R}}

\renewcommand{\strut}{\rule[-.3\baselineskip]{0pt}{1.5\baselineskip}}

\section{Methods}
\label{sect:method}
We first describe the neural dressing model that combines together deformable mesh modeling with neural rendering. After that we focus on the generative model for human avatars that builds on top of the neural dressing model.

\subsection{Neural dressing model} 

The neural dressing model (\fig{dressing}) builds on top of the deformable shape model (SMPL-X in our case) that generates fixed-topology mesh $M(\p,\s)$ driven by sets of pose parameters $\p$ and body shape parameters $\s$. We assume that the mesh comes with a pre-defined texture mapping function, and denote with $\Ren(M,T;C)$ the \textit{rasterization} function that takes mesh $M$, $L$-channel texture $T$ and camera parameters $C$, to generate $L$-channel rasterization of the textured mesh using z-buffer algorithm. 

We use $L$-channel texture $T$ with $L{=}16$ in our experiments to encode local photometric and geometric information (including the geometry missing in parametric mesh). We then use a rendering image-to-image network $f_\theta$ with learnable parameters $\theta$ to translate the $L$-channeled rasterized image $R$ to a six-channeled image $I$ of the same size, where the first three channels of $I$ correspond to RGB color channels and the last three channels correspond to segmentation masks for the whole foreground, head and hands. Additional segmentation channels for head and hands are used as additional supervision to improve visual quality of these body parts. 

In the neural dressing model, an avatar $A$ is characterized by body shape parameters $\s_A$ and neural texture $T_A$. Given a pretrained rendering network $f_\theta$, it can be rendered for an arbitrary pose $\p$ and arbitrary camera parameters $C$. At test time, the rendering process runs at interactive speed ($\sim$25 FPS at one megapixel resolution).

\subsection{Learning neural dressing via backpropagation} 
\label{sect:backprop}

% We then obtain 3-channel segmentation map -- with separate channel for the whole foreground, hands and head -- using pretrained network~\cite{gong2019graphonomy}
Given a collection of videos of several people, we can create their avatars by fitting our model via backpropagation. For person $i$, we assume that a set of video frames $I^j_i$ is given ($j \in 1..N_i$). We then obtain three-channel segmentation map using a pretrained network~\cite{gong2019graphonomy}, and fit parameters of body shape $\s_i$, pose $\p^j_i$ and camera $C^j_i$ corresponding to each individual frame. Our fitting is based on a modification of the  SMPLify-X algorithm~\cite{Pavlakos19} that constrains the body shape parameters to be shared between frames.

We then jointly optimize parameters of the rendering network $\theta$ and neural textures $T_i$ of all individuals using backpropagation. The optimization process minimizes following loss values: perceptual~\cite{Johnson16}, adversarial~\cite{Isola17}, and feature matching~\cite{Wang18} losses between ground truth images $I^j_i$ and rendered images $f_\theta[\Ren(M(\p^j_i,\s_i), T_i, C^j_i)]$. We use above-mentioned losses for color channels and Dice loss for ground truth and predicted segmentation masks. Importantly, when predicting the mask, we assign all pixels covered by the parametric body model to foreground. 

In the first set of experiments, we pretrain the rendering network parameters $\theta$ on a dataset of 56 individuals, for which we have collected many thousands of frames.
% We can then create an avatar for a new person , by taking fixed pretrained rendering network, estimating pose, body shape, and camera parameters for a limited number of frames of the individual, and then optimizing neural texture for the person by backpropagation of the same losses. 
We can then create an avatar for a new person in three steps: (i) estimate pose, body shape, and camera parameters for a limited number of frames of the individual, (ii) generate three-channel segmentation masks for those frames and then (iii) use the fixed pretrained rendering network to optimize neural texture for the person by backpropagation.
Note that in our current implementation, gradients are backpropagated through rasterization function only to texture parameters but not mesh or camera parameters, which means that non-differentiable renderers are suitable for our learning task.

\subsection{Generative modeling} \label{section_generative_modeling}

Learning with backpropagation allows to create avatars from relatively short videos or even few appropriately distributed frames (with each surface part of the person covered in at least one view). In some practical scenarios it may be beneficial to create fullbody avatars from impartial information, e.g.~a single view in A-pose. To do that, system needs ability to reason about unobserved parts of the avatar. We take the generative modeling approach to handle this task.

\begin{figure*}
    \centering
    \includegraphics[width=0.8\textwidth]{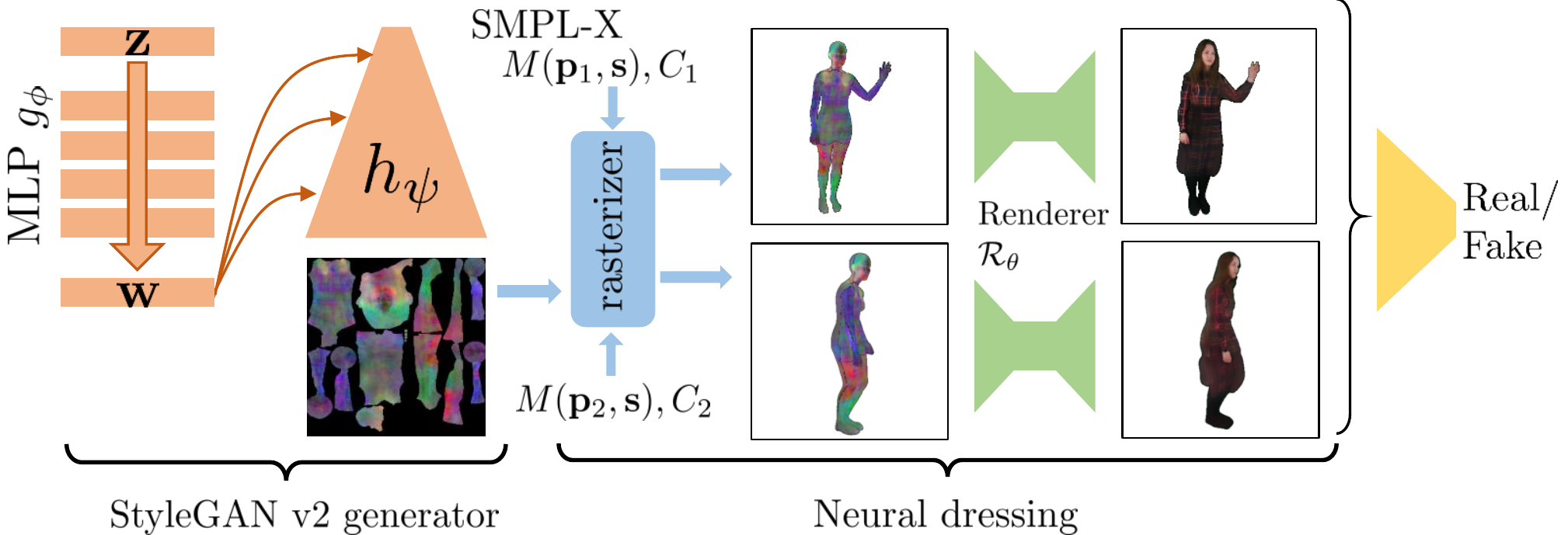}
    \caption{Our generative architecture is based on the combination of StyleGANv2 and Neural dressing. The StyleGAN part is used to generate neural textures, which are then superimposed on SMPL-X meshes and rendered with a neural renderer. During adversarial learning, the discriminator considers a pair of images of the same person. See text for more details.}
    \label{fig:architecture}
\end{figure*}

We build a generator for neural textures that can be used in neural dressing. We start with the recent StyleGANv2 approach~\cite{karras2019stylegan2} for high-resolution image modeling. Following~\cite{karras2019stylegan2}, we train a multi-layer perceptron $g_\phi(z): \R^{512} \to \R^{512}$ and a convolutional network $h_\psi(\w^4,...,\w^{512},\N)$ that takes as an input a set of $512$-dimensional \textit{style vectors} controlling generation at different resolutions (from $4\times{}4$ to $512\times{}512$) via modulation-demodulation mechanism as well as the set $\N$ of noise maps at these resolutions. Together, the two networks define the generative model 
% TODO: check notation, we need to either describe additional FC-layers for each scale (w -> w+) or regerd them as a part of generator h_\psi
% $h_\psi(g_\phi(z),g_\phi(z),\dots,g_\phi(z),\N)$,
$h_\psi(g_\phi(z),\N)$,
in which a random vector $\z$ is passed through the MLP and then used as an input at multiple resolutions into the convolutional part.

Our architecture for the generator (\fig{architecture}) closely follows~\cite{karras2019stylegan2} with two exceptions. First, in our case it outputs $L$-channeled neural texture instead of an RGB image. Second, we concatenate inputs of several later layers of the network with the 16-channel map of spectral coordinates of mesh vertices mapped into texture space to provide the generator with some information about the body mesh geometry.  During training, we also sample body poses and shapes from the training dataset, and superimpose the generated textures over the deformed body meshes. The resulting textured meshes are then passed through the rendering network that is trained as a part of the generator. Overall, the images produced by our generator are computed as follows:
\begin{gather} \label{eq:gen}
    % TODO: check notation
    % \Gen_{\phi,\psi,\theta}(\z,\N,\p,\s,C) = \\f_\theta\left[\Ren\left(\strut M(\p,\s),\; h_\psi(g_\phi(\z),g_\phi(\z),\dots,g_\phi(\z),\N),\; C\right)\right]\,,
    \Gen_{\phi,\psi,\theta}(\z,\N,\p,\s,C) = \\f_\theta\left[\Ren\left(\strut M(\p,\s),\; h_\psi(g_\phi(\z),\N),\; C\right)\right]\,,
\end{gather}
where $\z$ and $\N$ variables are sampled from the unit Normal distributions, while the pose, the camera, and the body shape variables are sampled from the empirical distribution in the training set.

Training of the full model is performed adversarially, with a dataset of videos of people as a source for real images. For each video we obtain foreground segmentation as well as body shape, pose and camera parameters as discussed above. 
%Adversarial training is driven by the discriminators. 
We utilize three different discriminator networks to train our system:
\begin{itemize}
    \item \textbf{Unary discriminator} closely follows StyleGANv2 architecture, and simply considers individual images (generated by our model as well as real images). The unary discriminator thus focuses on image quality. 
    \item \textbf{Binary discriminator} also follows StyleGANv2 architecture, however it takes pairs of images rather than individual images. Each real example is made up of two frames of a single video along with their segmentations. To obtain a fake example, we generate two examples $\Gen_{\phi,\psi,\theta}(\z,\N,\p_1,\s,C_1)$ and $\Gen_{\phi,\psi,\theta}(\z,\N,\p_2,\s,C_2)$, where $\z$, $\N$ (and thus the avatar texture), as well as the body shape $\s$ are shared between examples. Pose parameters $\p_1,\p_2$ and camera positions $C_1,C_2$ are different within each pair and correspond to two frames of the same video. The role of this discriminator is thus both to assess the visual realism of examples, as well as to ensure that the identity is preserved through camera change and pose change.
    \item \textbf{Face discriminator} considers individual images cropped around the face region (for both real and synthetic images) and is used to improve quality of models' face rendering. 
\end{itemize}

\paragraph{Additional regularization.} During training, we take additional steps to ensure identity preservation, as merely relying on binary discriminator turns out to be insufficient. We use three additional tricks. First, we train a predictor $q_\xi$ that takes generated images $I=\Gen_{\phi,\psi,\theta}(\z,\N,\p,\s,C)$ and tries to recover vector $\w=g_\phi(\z)$. The training loss $\|q_\xi(I)-\w\|^2$ is backpropagated through the entire generator, and thus ensures consistency of images created with same avatar texture.

The second regularizing trick is to ensure that rendering network is covariant to in-plane geometric transformation of its input. 
We achieve this by applying a random in-plane rigid transformation $\text{Tr}$ on rasterized images $J=\Ren(M(\p,\s),h_\psi(\z,\N),C)$. Transformed rasterizations $\text{Tr}(J)$ are then passed through the rendering network $r_\theta$, penalizing difference between $\text{Tr}(f_\theta[J])$ and $f_\theta[\text{Tr}(J)]$.
% This is ensured by taking rasterized images $J=\Ren(M(\p,\s),h_\psi(\z,\N),C)$ applying a random in-plane rigid transformation $\text{Tr}$, passing the original and the transformed inputs through the rendering network $r_\theta$, and then penalizing the difference between $\text{Tr}(f_\theta[J])$ and $f_\theta[\text{Tr}(J)]$.

Lastly, to mitigate the influence of poorly segmented images from our in-the-wild training dataset and those, where a part of a person's body is occluded, we force foreground masks of generated samples to cover the whole rendered mesh given as an input. To do that, instead of a sampled segmentation mask we use it's union with the binary mask of the mesh as a final foreground segmentation.

\subsection{Encoders and inference}

We now discuss how the generative model is used to fit a new avatar to one (or few) images of a person.
Given an image of the person $I$ (with corresponding segmentation mask), we estimate the parameters of its' body shape $\s$, pose $\p$, and camera $C$. Our goal is to find a texture $T$ such that the rendering $f_\theta[\Ren(M(\p,\s), T, C)]$ matches the observed image. The texture is parameterized by the convolutional generator $T=h_\psi(\w^4,\dots \w^{512},\N)$, and thus depends on the style vectors $\w^4,\dots \w^{512}$, the noise tensors $\N$ and the parameters $\psi$ of the generator $h_\psi$. 

\paragraph{Encoders.} As our texture parametrization is excessive (i.e.~the number of texture elements is usually greater than the number of observations in $I$), very different sets of latent variables are able to fit the observed image, while leading to different degree of generalization to new poses and camera parameters. We have therefore found it important to learn \textit{encoder networks} that would initialize latent vectors $\w$ at the part of the latent space that leads to good generalization. 

To learn the encoder network, we generate a dataset of synthetic samples from our texture generator. In particular, to get the $k$-th sample, we randomly and independently draw $\w_k^4,\dots \w_k^{512}$ by sampling $\z$ values and passing them through the perceptron part of the generative model. The noise tensors are drawn from the Normal distribution, and the convolutional generator produces a random texture. We then superimpose the texture on a random body from our training set in an approximate A-pose and pick an approximately frontal camera. The rendering network then produces an image $I_k$ of a random A-posed avatar. Our \textit{A-Encoder} $\Enc_A$ is then trained to recover the vectors $\w_k^4,\dots \w_k^{512}$ from image $I_k$. The training uses the L1-loss on synthetic data from our generative model only.

In addition to the A-encoder that is trained only on synthetic data, and is suitable for A-posed images, we have trained a generic encoder (\textit{G-encoder}) on both synthetic data and and pairs of video frames from real dataset. To train on real data the generic encoder $\Enc_G$ takes a pair of real images $(I_k,J_k)$ extracted from the same video of the same person, predicts latent variables $\w_k^4,\dots \w_k^{512}$ from $I_k$, augments them with random noise tensors $\N$ resulting in the texture  $T_k=h_\psi(\Enc_G(I_k),\N)$. The resulting texture is then superimposed on the mesh, posed and rendered according to the SMPL-X parameters $\p_k,\s_k$ and camera pose $C_k$ observed in the image $J_k$. Learned  perceptual  similarity (LPIPS)~\cite{zhang2018unreasonable} between $J_k$ and $f_\theta[\Ren(M(\p_k,\s_k), T_k, C_k)]$ is then used as a loss function for real data. For synthetic data, similarly to $\Enc_A$, we use the L1-loss between synthetic and predicted latent vectors. These two loss values are taken with equal weights resulting in the loss for the generic encoder $\Enc_G$.

The encoder's architecture is inspired by the pSp-architecture proposed in~\cite{richardson2020encoding}. The main idea is that each level of the encoder predicts a latent vector corresponding to the generator's resolution of this level. As a backbone we use EfficientNet-B7~\cite{tan2019efficientnet} pretrained on ImageNet~\cite{russakovsky2015imagenet}. More details about architecture is provided in Supplementary Material.

\paragraph{Fitting.} Avatar fitting to the given image(s) is done by optimization. The input to the optimization process is a set of images of a person $\boldsymbol{I} = \left\{ I_i | i \in [1, N]\right\}$. Latent vectors $\w$ are initialized by passing the images $\boldsymbol{I}$ through one of the two pretrained encoders. If more than one input image is available ($N>1$), the predicted latent vectors are averaged across all images. For the sake of simplicity we further assume that only one image $I$ is given as input. The optimization is performed over (i) the latent vectors $\w$, (ii) the parameters $\psi$ of generator $h_\psi$ and (iii) the noise tensors $\N$ to further minimize the difference between the recovered image $f_\theta[\Ren(M(\p,\s), T, C)]$ and $I$. As a last stage of the optimization, we perform direct optimization of the texture values for 100 iterations.

During the optimization we use multiple loss functions, namely LPIPS-loss~\cite{zhang2018unreasonable} between recovered image and input image, Mean Squared Error (MSE) between recovered image and input image, Mean Absolute Error (MAE) on latent variables $w$ deviation from the initialization predicted by the encoder,  MAE on generator parameters $h_\psi$ deviation from the initial ones, MAE on texture deviation from the texture values, optimized to the beginning of the last stage, LPIPS-loss~\cite{zhang2018unreasonable} on face regions of recovered and input images, and feature matching loss based on the trained face discriminator~\cite{Pan20}. In case of multiple input images, losses are averaged across all images. We provide additional analysis on the role of losses in the Supplementary Material. Optimization is performed via backpropagation with ADAM optimizer~\cite{kingma2014adam}.

\section{Experiments}
\label{sect:experiments}
Below, we present experimental validation of the neural dressing approach. We first evaluate the approach in the scenario, where an avatar is created from a short video of a person, using backpropagation without generative prior on the neural texture. We then proceed to evaluate samples from our generative model and its performance for one-shot and few-shot image-based full-body avatar creation. We provide additional results including qualitative comparisons in the Supplementary material.

\paragraph{Datasets.} 
We use our recently collected \textit{AzurePeople} dataset to pretrain rendering network in backpropagation-based approach. It was collected with five Kinect Azure sensors and contains multi-view RGBD-videos of 56 people (though we do not use depth information in this work).

Four other datasets were used to train generative model. The first is a multi-view \textit{HUMBI} dataset~\cite{yu2020humbi}. At the moment of experiments it has 442 subjects available with total of 61,975 images. This dataset alone was used to train binary discriminator and ensure multi-view consistency of generated samples. The second dataset \textit{TEDXPeople} was collected by us and contains 41,233 videos of TED and TED-X talks. For our purposes we have extracted and processed 8 to 16 frames from each video, in which a person is visible in either full-body or close-up upper-body view. This dataset is very diverse in terms of demographics and clothing styles, though the resolution is limited and there are certain biases associated with peculiar lighting and camera viewpoints specific to TED(-X) talks. Because of these limitatios, we only used this dataset for the unary discriminator.

Additionally, we used the dataset containing \textit{curated SMPL-X fits} released by the authors of ExPose~\cite{ExPose:2020} and the \textit{Google OpenImages v6} dataset~\cite{OpenImages} (from which we extracted human bounding boxes). After filtering out low-resolution and highly occluded samples from these two datasets, we added remaining 10869 \textit{ExPose} samples and 37660 \textit{OpenImages} samples to our training procedure.

For evaluation only, we also use two videos from the PeopleSnapshot dataset~\cite{alldieck2018video} ('female-1-casual' and 'male-2-casual'). Each video has a person  rotating in front of the camera in A-pose. We choose these particular sequences because we have access to their avatars created with several previous methods.

\subsection{Video-based avatars}

\begin{figure*}
    \centering
    \includegraphics[width=0.85\textwidth]{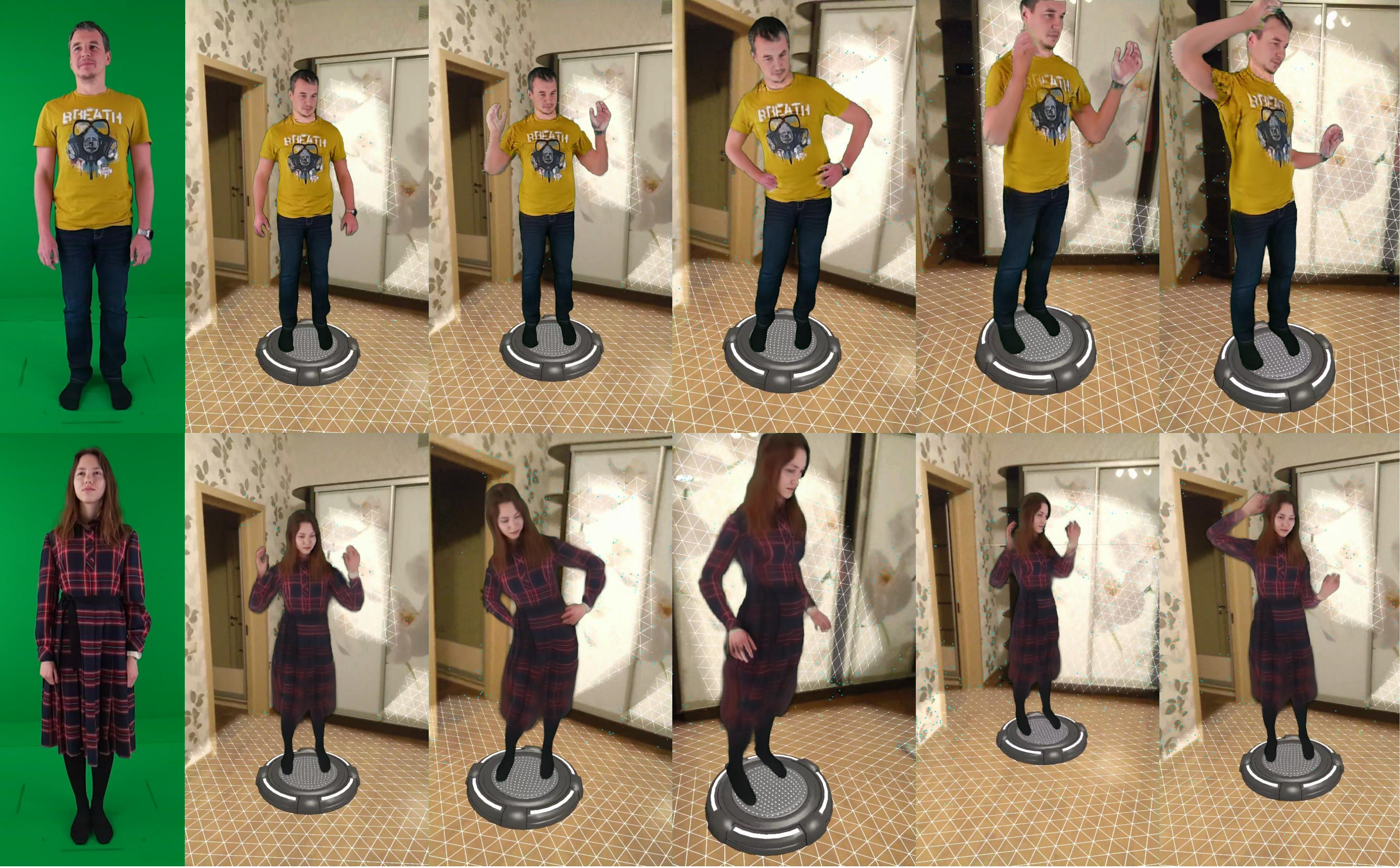}
    \caption{Results for video-based avatars. Sample source frames are shown on the left. The remaining images show the video-based avatars in previously unseen poses shown within a simple augmented reality application. The pose sequence is transferred from a different person.}
    \label{fig:azure}
\end{figure*}

In the first set of experiments, we assess the ability of our models to create avatars from videos, in which one person is seen in different poses. Here, we create avatars using backpropagation as described in \sect{backprop}. We show qualitative results of this approach on people from AzurePeople dataset in \fig{azure} and in supplementary video. 

For evaluation, we also compare our results to textured neural avatar system~\cite{shysheya2019textured}, and mesh-based approach~\cite{alldieck2018video}. Since we do not have any hold-out video sequences for the models from \cite{alldieck2018video}, we perform a user study using an online crowd-sourcing platform showing short videos of two people from PeopleSnapshot dataset in pose sequences sampled from TEDXPeople dataset. For visual quality, participants preferred our method in 69.8\% of the cases when compared to~\cite{shysheya2019textured} and in 70.6\% of the cases when compared to~\cite{alldieck2018video}. For identity preservation, preferences were 67.7\% and 69.8\% respectively.

We have additionally compared our method to~\cite{shysheya2019textured} on in-the-wild YouTube video considered in~\cite{shysheya2019textured}. It contains much more challenging poses than TEDXPeople samples. Here participants preferred our method 63.9\% of the cases when asked about visual realism and in 52.6\% of the cases when asked about identity preservation.

In both studies participants were presented with two-second segments of 15 fps videos (10 segments for PeopleSnapshot sequences and 18 for the YouTube sequence) and asked to choose from pairs of side-by-side positioned samples. In identity preservation study, an image from ground truth video was presented along with the compared results. The order of compared results for each pair was randomized. Each pair was assessed by 30 people.

\subsection{Few-shot avatars}

\begin{figure*}
    \centering
    \includegraphics[width=0.9\textwidth]{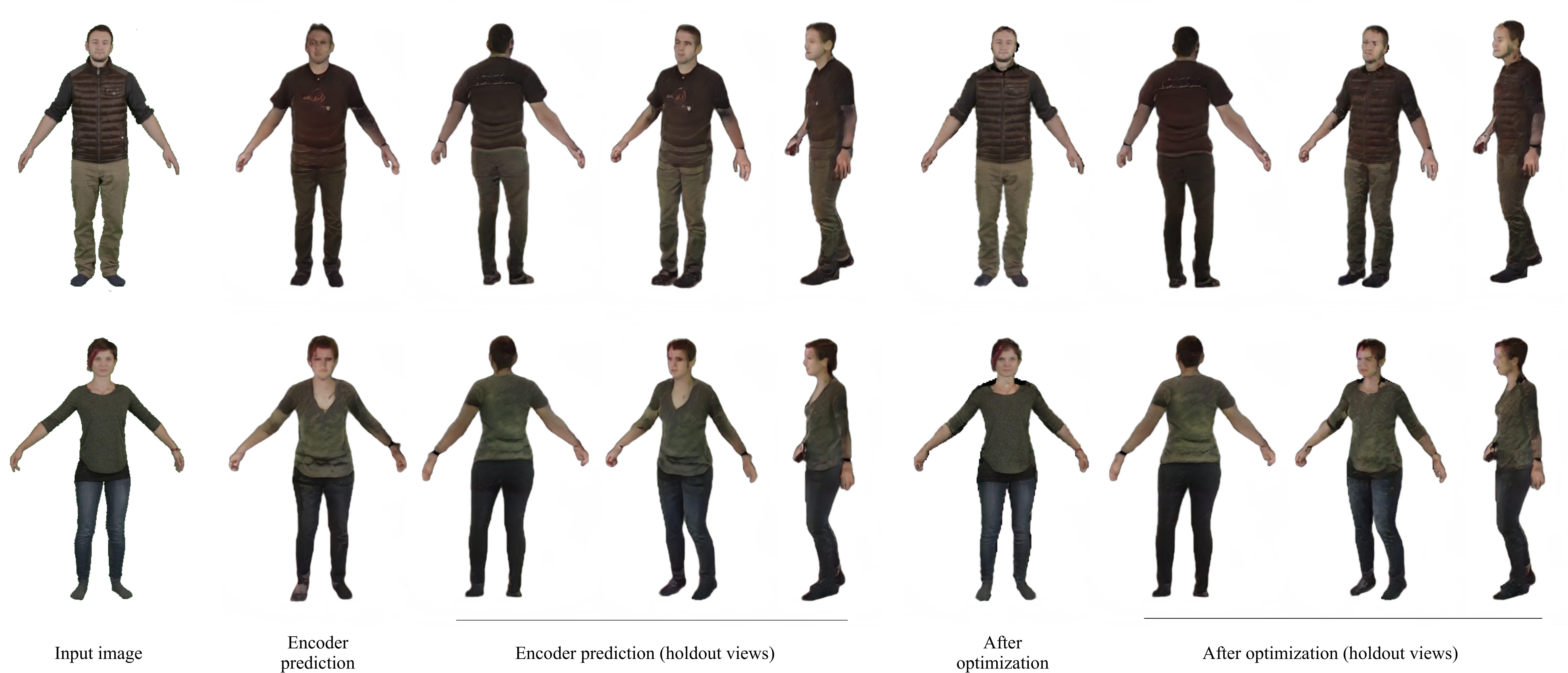}
    \caption{One-shot inference results for test subjects from People Snapshot dataset~\cite{alldieck2018video}. Single \textit{Input image} is fed into the encoder to produce initial texture estimation (\textit{Encoder prediction}). Next, optimization step draws output renderings closer to the original person (\textit{After optimization}). Note, how non-visible details in the source image  propagate to the output from the initial encoder prediction (e.g.~the back of the male's shirt).
    }
    \label{fig:oneshot}
\end{figure*}

We then focus our evaluation on the few-shot scenario. We first evaluate our avatar against the mesh-based Octopus approach~\cite{alldieck2019learning} in the eight-shot mode (i.e.\ each avatar is created from eight photographs). For one-shot mode, we compare against the mesh-based method~\cite{lazova2019360}, as well as three neural rendering/warping one-shot methods: the first-order motion model~\cite{siarohin2019first}, the liquidGAN model~\cite{liu2019liquid}, and the coordinate-based inpainting model~\cite{grigorev2018coordinate}. The first-order motion model was retrained on the TEDXPeople dataset (for completeness we also report the results for the author's model trained on the Taichi dataset). For other methods, author-provided pretrained models/results were used (see the supplementary material for details). For our method, we consider two variants initialized with two different encoders. \fig{oneshot} demonstrates process of fitting our models to the data (using the A-encoder).

We use a set of standard metrics to evaluate image quality of the avatars, including learned perceptual similarity (LPIPS)~\cite{zhang2018unreasonable}, structured self-similarity (SSIM)~\cite{Wang04}, Fréchet Inception Distance (FID)~\cite{Heusel17}, and Inception Score (IS)~\cite{Salimans16}. We use two-person People Snapshot dataset to compare with~\cite{alldieck2019learning},~\cite{lazova2019360},~\cite{liu2019liquid} and~\cite{siarohin2019first}. (Table ~\ref{table:people_snapshot_few_shot}) as those were available to us for all methods. We also compare to the state-of-the-art one-shot methods on three hold-out video sequences from test split of our TEDXPeople dataset (Table~\ref{table:tedx_few_shot},~\fig{oneshot}). Both comparisons show advantage of our approach in all metrics, except the Inception Score, which we subjectively find least correlated with visual quality. Qualitative results and additional comparisons are presented in the supplementary material and video.

It has to be mentioned, that while all one-shot models create avatars from the same data (a single image), our model requires a lengthy optimization process to come up with an avatar, which is a limitation of our model.

% \begin{figure*}
%     \centering
%     \includegraphics[width=\textwidth]{figures/dress8.pdf}
%     \caption{Results in A-pose for eight-shot mode for one of the two test avatars from the PeopleSnapshot dataset. Left three images -- the Octopus system~\cite{alldieck2019learning}, right four images -- ours. Digital zoom-in recommended. Our approach results in a more realistic avatar. }
%     \label{fig:render_people}
% \end{figure*}

% \begin{itemize}
%     \item FID/IS on pose sequence from tedx
%     \item FID/IS/LPIPS/SSIM/user study on rotated A-pose
%     \item demo for a rotated A-pose (vs \cite{lazova2019360}) / sequence of tedx poses (vs \cite{alldieck2019learning})
% \end{itemize}

\begin{table}[h!]
\centering
\resizebox{0.45\textwidth}{!}{
 \begin{tabular}{||c c c c c||} 
 \hline
  & IS$\uparrow$ & FID$\downarrow$ & LPIPS$\downarrow$ & SSIM$\uparrow$ \\
 \hline\hline
 \multicolumn{5}{||c||}{One-shot} \\
  \hline
  360Degree~\cite{lazova2019360} & \textbf{1.8643} & 1383.1 & 0.2123 & 0.8079 \\
%   \hline
%   CBI~\cite{grigorev2018coordinate} & - & - & - & - \\
  \hline
  LWGAN~\cite{liu2019liquid} & 1.7159 & 1771.9 & 0.2727 & 0.2876 \\
  \hline
  FOMM-TaiChi~\cite{siarohin2019first} & 1.7643 & 1059.4 & 0.1767 & 0.7585 \\
  \hline
  FOMM-TEDX~\cite{siarohin2019first} & 1.5404 & 674.3 & 0.1477 & 0.8743 \\
  \hline

%   Neural Dressing (A-encoder) & 1.8266 & 367.7 & \textbf{0.0826} & \textbf{0.9090} \\
  Neural Dressing (A-encoder) & 1.8025 & 349.2 & 0.1957 & 0.8264 \\
  \hline
%   Neural Dressing (G-encoder)  & \textbf{1.8594} & \textbf{332.2} & 0.0842 & 0.9065 \\
    Neural Dressing (G-encoder) & 1.7469 & \textbf{272.1} & \textbf{0.0836} & \textbf{0.9012} \\
  \hline
%  Supervised Neural Dressing & 1.8496 & 246.1 & 0.0783 & 0.9069 \\
  \hline
  \multicolumn{5}{||c||}{8-shot} \\
  \hline
%  Octopus (bad align) & 1.6106 & 412.98, & 0.16948 & 0.85906 \\ 
%  \hline
 Octopus~\cite{alldieck2019learning} & \textbf{1.8123} & 403.6 & 0.1379 & 0.8324 \\ 
%  \hline
%  Neural Dressing (old) & \textbf{1.8586} & \textbf{153.8} & \textbf{0.0696} & 0.8925 \\
  \hline
%  Neural Dressing (A-encoder) & \textbf{1.7103} & 260.7 & \textbf{0.0708} & \textbf{0.9155} \\
 Neural Dressing (A-encoder) & 1.7335 & 183.3 & 0.1169 & 0.8764 \\
   \hline
%  Neural Dressing (G-encoder) & 1.5997 & \textbf{223.2} & 0.0714 & \textbf{0.9155} \\
 Neural Dressing (G-encoder) & 1.6608 & \textbf{158.2} & \textbf{0.0759} & \textbf{0.9079} \\
 \hline
 %Supervised Neural Dressing & 1.6551 & 119.1 & 0.0536 & 0.9280 \\
   
 \hline
\end{tabular}
}
\caption{Comparison against state-of-the-art few-shot methods on two sequences from People Snapshot dataset~\cite{alldieck2018video}. We use the first frame as a source image for one-shot methods, and the frames used in ~\cite{alldieck2019learning} for eight-shot methods, and evaluate on the remaining frames of the sequences. Metrics are calculated with respect to the ground truth sequence.}
\label{table:people_snapshot_few_shot}
\end{table}

% \begin{figure*}
%     \centering
%     \includegraphics[width=\textwidth]{figures/dress_1.pdf}
%     \caption{The comparison of one-shot fullbody rerendering methods. Given the source image, the goal of each of the compared methods is to reproduce the ground truth image. While our method returns a reasonable results, the methods we compare to (coordinate-based inpainting, retrained FOMM, Liquid GAN) are not capable to handle the task well.}
%     \label{fig:oneshot}
% \end{figure*}

\begin{table}[h!]
\centering
\resizebox{0.45\textwidth}{!}{
 \begin{tabular}{||c c c c c||} 
 \hline
  & IS$\uparrow$ & FID$\downarrow$ & LPIPS$\downarrow$ & SSIM$\uparrow$ \\
 \hline
 \hline
  \multicolumn{5}{||c||}{One-shot} \\
  \hline
  
  CBI~\cite{grigorev2018coordinate} & 1.4738 & 6761.3 & 0.1814 & 0.8293 \\
  \hline
  LWGAN~\cite{liu2019liquid} & \textbf{1.7467} & 10671.9 & 0.3284 & 0.6068 \\
  \hline
  FOMM-TaiChi~\cite{siarohin2018deformable} & 1.2981 & 3253.1 & 0.2186 & 0.7725 \\
  \hline
  FOMM-TEDX~\cite{siarohin2018deformable} & 1.2663 & 3976.1 & 0.2312 & 0.7944 \\
  \hline
   %Neural Dressing (A-encoder) & 1.3439 & \textbf{1410.6} & \textbf{0.1020} & \textbf{0.8745} \\
   Neural Dressing (A-encoder) & 1.4448 & \textbf{601.7} & \textbf{0.0531} & \textbf{0.9369}  \\
   
   \hline
   %Neural Dressing (G-encoder) & \textbf{1.4655} & 1760.5 & 0.1056 & 0.8723 \\
   Neural Dressing (G-encoder) & 1.5903 & 765.7 & 0.0542 & 0.9361 \\

  \hline
  \multicolumn{5}{||c||}{8-shot} \\
    \hline
  %Neural Dressing (A-encoder)  & 1.4226 & \textbf{1349.7} & \textbf{0.0983} & \textbf{0.8784} \\
  Neural Dressing (A-encoder)  & 1.4372 & \textbf{505.2} & \textbf{0.0521} & \textbf{0.9389} \\
  \hline
  %Neural Dressing (G-encoder)  & \textbf{1.4892} & 1594.5 & 0.1005 & 0.8765 \\
  Neural Dressing (G-encoder)  & \textbf{1.5548} & 559.8 & 0.0537 & 0.9381 \\
  \hline
\end{tabular}
}
\caption{Comparison against state-of-the-art one-shot methods on three test sequences of the TEDXPeople dataset. Our method performs better than competitors in most metrics.}
\label{table:tedx_few_shot}
\end{table}

\subsection{Generative model ablation}
Additionally, we evaluate contributions of regularization mechanisms and additional discriminators in the training process of generative model. We compare ablations in terms of two metrics. First, we use Fréchet Inception Distance (FID) between the set of real validation images and the set of sampled images of people with the same meshes as in the validation set. The second metric is the Euclidean distance between vectors extracted with a re-identification network~\cite{zhou2019learning} from two images rendered with same neural texture, but positioned with pose difference of $180^{\circ}$ (denoted as ReID). While the first metric focuses on the image quality, the second measures consistency of rendering.

\begin{table}[h!]
\small
\centering

\resizebox{0.25\textwidth}{!}{
\begin{tabular}{||l l l ||}
\hline
                                        & FID$\downarrow$           & ReID$\downarrow$ \\ \hline 
\hline
\textit{Ours\_full}                     & 48.3          & \textbf{15.0} \\ \hline
\textit{no-augmentations}     & 48.8          & 16.3 \\ \hline
\textit{no-meshmask}   & 54.2          & 15.8 \\ \hline
\textit{no-speccoords} & 47.8          & 15.5 \\ \hline
\textit{no-latent-predictor}       & 44.2          & 15.9 \\ \hline
\textit{no-face-discr}    & 43.4          & 16.0 \\ \hline
\textit{no-bin-discr}     & \textbf{42.9} & 17.3 \\ \hline
\end{tabular}
}
\caption{Ablation study for generative model of neural textures}
\label{table:gen_ablation}
\end{table}

With each subsequent ablation we turn off one of the modifications of the training procedure. Namely, \textit{no-augmentations} loses in-plane rotation regularization \textit{no-meshmask} uses plain sampled segmentation as foreground mask instead of its' union with the mask of SMPL-X mesh projection, \textit{no-speccoords} omits spectral coordinates passed into the texture generator, \textit{no-latent-predictor} does not use latent predictor $q_\xi$, \textit{no-face-discr} and \textit{no-bin-discr} omit facial and binary discriminator respectively.

As seen from Table~\ref{table:gen_ablation}, although the most basic variant provides lowest FID for single images, the modifications improve consistency across views.

%Model with all contributions enabled gets the FID 10.73. Turning of the in-plane geometric transform regularization leads to the increase of FID to 11.24. Cutting off the loss signal from the $q_\xi$ network, which tries to recover the latent vector $\w=g_\phi(\z)$ from generated image and segmentation, additionally increases FID to 12.20. Replacing the pair discriminator (described in Section~\ref{section_generative_modeling}) with the standard ``unary'' discriminator that takes a single sample from the generator (4-channel input) leads to FID 13.16. Thus, each of the above-mentioned improvements appears to be important.

% retrain~\cite{siarohin2019first} on our TEDX dataset and compare on 
% \begin{itemize}
%     \item test images from TEDX dataset (FID/IS/LPIPS/SSIM/user study)
%     \item rotated pose from TEDX dataset (FID/IS/user study)
%     \item  demo video on tedx pose sequences
% \end{itemize}

% We compare to~\cite{liu2019liquid}, \cite{grigorev2018coordinate}
% \begin{itemize}
%     \item DeepFashion (image pairs, FID/IS/LPIPS/SSIM/user study)
%     \item rotated pose (FID/IS/user study)
%     \item demo video on tedx pose sequences
% \end{itemize}

% \subsection{Ablations}
% \begin{itemize}
%     \item 
% \end{itemize}

\section{Conclusion}
\label{sect:conclusion}

We have presented a new approach for modeling the human appearance. This approach integrates polygonal body mesh modeling with neural rendering. Thus, polygonal mesh controls and models coarse body geometry as well as the pose, while neural rendering component adds clothing and hair. On top of that, we train a generative model for neural textures, and show that this generative model allows to obtain avatars in few-shot mode. We show that proposed approach improves state-of-the-art for video-based and few image-based avatar acquisition. Admittedly, results of our method are still limited by the amount and quality of data, especially diversity the biases of video sequences. Improving data efficiency of our generative modeling is thus one of the important directions of out future work.

\newpage
\setcounter{section}{1}
\section*{A. Neural dressing}
\subsection*{Model details}
The Neural Dressing model consists of two elements: the neural texture and the renderer network.

\textbf{The neural texture} is a three-dimensional $512\times{}512\times{}16$ tensor. During training we store it as an array of severn mipmaps (from 8p to 512p), which allows us to regularize high-resolution mipmaps and encourage low-resolution ones to store more information (as was proposed in~\cite{thies2019deferred}).

\textbf{The renderer network} $R$ is a standard U-net model from~\cite{Yakubovskiy2019} with Resnet-16 encoder backbone. This network processes the neural texture warped onto the SMPL-X mesh along with additional two channels containing mesh with UV texture coordinates. The main part of the network outputs a 16-channel tensor which is then passed through two parallel shallow convolutional networks. The first of these networks generates the three-channel RGB image, while the second produces a three-channel segmentation mask. The first channel of the segmentation is the foreground mask, while the second and the third are hands and face masks respectively (which are used for additional supervision).

\section*{B. Supervision for video-based training}
As we train our Neural Dressing model on high-resolution videos from \textit{AzurePeople} dataset, it allows us to use several supervisory signals. Apart from standard \textbf{VGG19 perceptual loss} $\mathcal{L}_{vgg}$ between the output images and the ground truth frames, several other loss functions were used.

We have obtained three-channel segmentation maps (as described above) using Graphonomy network~\cite{gong2019graphonomy} and then refined it with Photoshop's chromakey feature.
These segmentation maps $M_3$ are used to enforce rendering network to generate realistic foreground masks as well as to learn to distinquish face and hands regions for better visual quality of these regions. For segmentation supervision, \textbf{Dice loss} is used:
% \begin{gather} \label{eq:gen}
%     \Gen_{\phi,\psi,\theta}(\z,\N,\p,\s,C) = \\f_\theta\left[\Ren\left(\strut M(\p,\s),\; h_\psi(g_\phi(\z),g_\phi(\z),\dots,g_\phi(\z),\N),\; C\right)\right]\,,
% \end{gather}
$$
\mathcal{L}_{segm} = -\log{\frac{2\|M_3 \odot \hat{M}_3 \|}{\|M_3\| + \| \hat{M}_3 \|}},
$$
where $\hat{M}_3$ is the predicted segmentation and $\odot$ is Hadamard product.

To ensure realistic high-frequency details, a set of discriminators was used. The first discriminator distinguishes between real and fake samples of full images. The other two discriminators are given image crops around face and head regions respectively. The crops are calculated using OpenPose keypoint detector~\cite{cao2018openpose}. The cropped regions are also masked by respective channels in segmentation tensors. All discriminators employ PatchGAN structure with $l2$ adversarial losses:
% $$

\begin{gather*}
    \mathcal{L}_\text{adv}^R = \|1 - D_\text{full}(\hat{I})\|_2  \\
    \mathcal{L}_\text{adv}^D = \|D_\text{full}(\hat{I})\|_2 + \|1 - D_\text{full}(I)\|_2 
\end{gather*}

\begin{gather*}
    \mathcal{L}_\text{advface}^R = \|1 - D_\text{head}(\mathcal{C}_\text{head}(\hat{I}\odot \hat{M}_\text{head}))\|_2  \\
    \mathcal{L}_\text{advface}^D = \|D_\text{head}(\mathcal{C}_\text{head}(\hat{I}\odot \hat{M}_\text{head}))\|_2 + \\ \|1 - D_\text{head}(\mathcal{C}_\text{head}(I\odot M_\text{head})\|_2 
\end{gather*}

\begin{gather*}
    \mathcal{L}_\text{advhands}^R = \|1 - D_\text{hands}(\mathcal{C}_\text{hands}(\hat{I}\odot \hat{M}_\text{hands}))\|_2  \\
    \mathcal{L}_\text{advhands}^D = \|D_\text{hands}(\mathcal{C}_\text{hands}(\hat{I}\odot \hat{M}_\text{hands}))\|_2 + \\ \|1 - D_\text{hands}(\mathcal{C}_\text{hands}(I\odot M_\text{hands})\|_2
\end{gather*}

In the equations above, $M_\text{hands}$ and $M_\text{head}$ are the segmentation masks for hands and head regions respectively, ($\hat{M}_\text{hands}$ and $\hat{M}_\text{hands}$ correspond to the generated masks). $\mathcal{C}_\text{hands}$ and $\mathcal{C}_\text{head}$ are crop operators for hands and head regions. For hands, the crop for each hand is passed through the hand discriminator separately. $I$ and $\hat{I}$ denote full ground truth and generated images respectively. To enhance face realism we also employ the \textbf{VGGFace perceptual loss} $\mathcal{L}_\text{VGG Face}$.

The last loss term regularizes the high-resolution texture mipmaps so that they store only high-frequency details:
\begin{gather*}
    \mathcal{L}_\text{mipmap} = \sum_{i=3}^9 \alpha_{i}\|T^{(2^i)}\|_2 ,
\end{gather*}
where $T^{(2^i)}$ is a mipmap of resolution $2^i$, and $\alpha_i$ has the following values: $\underset{i=\overline{3,9}}{\alpha} = \{0,0,0,1,2,4,8\}$

\section*{C. Training procedure}
In the video-based setup, the model is trained in two phases. First, the renderer network is pretrained on all 56 people from \textit{AzurePeople} dataset along with the corresponding neural textures. In this phase, the renderer is learning to convert 512p neural renders into 512p images.

\begin{table}[h!]
\centering
\resizebox{0.31\textwidth}{!}{
\begin{tabular}{||c|c|c||}
\hline
                    & phase 1 & phase 2 \\  \hline\hline
\multicolumn{3}{||c||}{loss weights}                        \\ \hline
$\mathcal{L}_{vgg}$                 & 1                & 1                \\ \hline
$\mathcal{L}_{vggface}$             & 2e-1             & 2e-1             \\ \hline
$\mathcal{L}_{segm}$                & 1e+2             & 1e+2             \\ \hline
$\mathcal{L}_{adv}$                 & 1e+1             & 1e+1             \\ \hline
$\mathcal{L}_{advface}$             & 5e+1             & 5e+1             \\ \hline
$\mathcal{L}_{advhands}$            & 5e+1             & 5e+1             \\ \hline
$\mathcal{L}_{mipmap}$            & 1             & 1             \\ \hline
\multicolumn{3}{||c||}{learning rates}                      \\ \hline
Texture             & 5e-2             & 1e-3             \\ \hline
$R$                   & 1e-3             & 1e-3             \\ \hline
$\mathcal{D}_{full}$               & 2e-4             & 2e-4             \\ \hline
$\mathcal{D}_{head}$              & 2e-4             & 2e-4             \\ \hline
$\mathcal{D}_{hands}$             & 2e-4             & 2e-4             \\ \hline
\multicolumn{3}{||c||}{other parameters}                    \\ \hline
$R$ optim beta1       & 5e-1             & 5e-1             \\ \hline
Texture optim beta1 & 5e-1             & 5e-1             \\ \hline
\end{tabular}
}
\caption{Hyperparameters for both phases of video-based neural dressing training procedure}
\label{table:neuraldressing_hyperparameters}
\end{table}

In the second step, the network is finetuned for a single person (either from the same dataset or completely new one) at 1024p resolution, while the maximum resolution of neural texture mipmaps remains 512p. During both steps ADAM optimiser is used for all networks. Table~\ref{table:neuraldressing_hyperparameters} shows the list of hyperparameters used during both phases.

\section*{D. Generative Textures}
\subsection*{Architecture details}
Here we describe modifications to the Neural Dressing approach that allow us to sample human body neural texture from a generative model. In this setup we still have the rendering network $R$, while the stack of neural texture mipmaps is substituted by the generative network $G$.

The architecture of $G$ closely follows the StyleGANv2 model~\cite{karras2019stylegan2}. The only modifications of the architecture of~\cite{karras2019stylegan2} are additional inputs to several convolutional layers of the network. We pass additional 16 channels of SMPL-X mesh vertices spectral coordinates alongside the outputs of previous layer. This modification affects layers receiving 64p, 128p and 256p feature maps,  To acquire these maps, we calculate spectral coordinates for each SMPL-X mesh vertex and then rasterize them bilinearly in the UV texture space.

The generator $G$ outputs 16-channel 256p feature maps, which are then warped onto the SMPL-X UV render and passed through the renderer $R$, which outputs three-channel RGB output along with one-channel foreground segmentation. The latter is then modified so that at every pixel on which the SMPL-X is projected,  the value is set to one (i.e.~the projection of the SMPL-X mesh is forced to belong to the foreground). This enforcement increases the stability of segmentation.

Apart from $R$ and $G$, we use three discriminator networks: $D_\text{unary}$, $D_\text{binary}$ and $D_\text{face}$. All of them follows the discriminator architecture from~\cite{karras2019stylegan2}.

$D_\text{unary}$ takes four-channel image (RGB+foreground segmentation) as an input. $D_\text{binary}$ takes two such images of the same person in different poses stacked. $D_\text{face}$ is given a four-channel 128p crop of the face region. 

\subsection*{Loss functions}
For our adversarial training procedure we use the loss from~\cite{karras2019stylegan2}, adopting same non-saturating loss for discriminator outputs $\mathcal{L}_{adv\_unary}$, $\mathcal{L}_{adv\_binary}$ and $\mathcal{L}_{adv\_face}$ and the same regularization techniques, i.e.~R1 regularization for discriminators $\mathcal{L}_{r1}$ and path regularization for generator $\mathcal{L}_{path}$.

We also add two new regularizations specific to our system described in the \textbf{Additional regularization} paragraph of the main paper.

\begin{gather*}
    \mathcal{L}_{wreg} = \|q_\xi(I)-\w\|^2 \\
    \mathcal{L}_{augreg} = \| \text{Tr}(f_\theta[J]) - f_\theta[\text{Tr}(J)] \|
\end{gather*}

\subsection*{Training procedure}
The training of the generative model is also split in two phases. First, we train the pipeline to produce 256p images and then tune them for the 512p setup, while filtering out low resolution images from the training dataset. In both phases, the generator $G$ has the same set of layers and outputs 256p neural textures, while the renderer $R$ is given an additional upsampling and several convolutional layers during the switch of the phases. Therefore, the final version of $R$ takes 256p neural renders and outputs 512p images. The discriminators $D_{unary}$ and $D_{binary}$ are also augmented with additional layers in the beginning so that they are able to process 512p inputs. All additional layers are gradually (progressively) introduced into the training procedure following~\cite{Karras18}.

\begin{table}[h!]
\centering
\resizebox{0.18\textwidth}{!}{
\begin{tabular}{||c|c||}
\hline
\multicolumn{2}{||c||}{loss weights}                        \\ \hline
$\mathcal{L}_{adv\_unary}$  & 1              \\ \hline
$\mathcal{L}_{adv\_binary}$ & 1              \\ \hline
$\mathcal{L}_{adv\_face}$   & 1              \\ \hline
$\mathcal{L}_{r1}$        & 1e+1           \\ \hline
$\mathcal{L}_{path}$    & 2              \\ \hline
$\mathcal{L}_{augreg}$     & 1              \\ \hline
$ \mathcal{L}_{wreg}$        & 1e+1           \\ \hline
\multicolumn{2}{||c||}{learning rates}                        \\ \hline
$G$           & 1e-3           \\ \hline
$R$           & 1e-4           \\ \hline
$q_\xi$           & 1e-3           \\ \hline
$D_{unary}$     & 2e-3           \\ \hline
$D_{binary}$   & 2e-3           \\ \hline
$D_{face}$      & 2e-3           \\ \hline
\end{tabular}
}
\caption{Hyperparameters for generative texture model training}
\label{table:generative_hyperparameters}
\end{table}

For all networks in the pipeline we also employ equalized learning rate technique introduced in~\cite{Karras18}. Training hyperparameters are the same for both phases and are listed in Table~\ref{table:generative_hyperparameters}

% \section{Training details}
% We train our model in two phases. During the first one, both texture generation and neural rendering parts of a generative model are trained along with a discriminator and latent variable predictor $q_\xi$. At the start of the second phase, we add the face discriminator to enhance quality of human faces in our samples. For both phases we use ADAM optimizer with the batch of eight samples and run the training on eight NVidia Tesla P100 GPUs. We make 100000 optimizing steps in the first phase and 80000 steps in the second.

% Following the training procedure of StyleGANv2~\cite{karras2019stylegan2} we update the weights of the generator and the discriminators in alternating manner. Every ten training iterations we also do a separate update only for the neural rendering part of our generator, in which we only propagate regularization loss enforcing equivariance to in-plane geometric transformation.

% We also find that lowering learning rate for neural rendering part forces it to perceive only local information in the sampled neural texture, which ensures that no person-specific information is encoded within the SMPL silhouette shape and therefore allows for more diverse texture samples. We therefore optimize the neural renderer with
% the learning rate $2e^{-4}$, while all other parts of the pipeline are optimized with the learning rate $2e^{-3}$.

\section*{E. Details about other methods}

We list the methods we compare against, and provide details about the (re)-implementations.

{\bf Textured Neural Avatars~\cite{shysheya2019textured}.} We use the models provided by the authors.

{\bf Videoavatars~\cite{alldieck2018video}} We use the models provided by the authors.

{\bf 360 degree~\cite{lazova2019360}}. Since the authors provide only a non-animatable mesh of a person in the A-pose, we have rendered the mesh under a visually close global rotation  and then aligned the image with ground truth using an affine warp computed from the OpenPose detections.

{\bf Octopus~\cite{alldieck2019learning}}. We use the 3D model provided by the authors. We have rendered the images in the SMPL poses provided with the dataset. However due to scale mismatch the alignment with ground truth was poor, and we therefore applied affine warping based on OpenPose as for the previous baseline.

{\bf LWGAN~\cite{liu2019liquid}} We use the network provided by the authors. Since the method does not produce a segmentation mask, we have replaced the color of the background pixels with white in the input images in order to maintain consistency in the comparison.

{\bf FOMM~\cite{siarohin2019first}.} We use the networks provided by the authors that was trained on TaiChi dataset (we refer to this baseline as FOMM-TaiChi). Since our TEDX dataset differs significantly from TaiChi (stage performances vs. martial arts), we also train FOMM from scratch on the TEDX data (the corresponding baseline is called FOMM-TEDX). FOMM-TEDX baseline is trained on $41233$ sequences each containing eight images.

\textbf{Qualitative comparison to PIFu and PIFuHD.} In addition to the comparisons in the main text, we compare qualitatively to PIFu~\cite{saito2019pifu} and PIFuHD~\cite{saito2020pifuhd}. We use the networks provided by the authors. We use them to create 3D models by one frontal image from PeopleSnapshot dataset and render them from new viewpoint for comparison. Since PIFuHD model does not produce full-body texture, we use normals of the 3D model for its' visualization.

\begin{figure*}
    \centering
    \includegraphics[width=\textwidth]{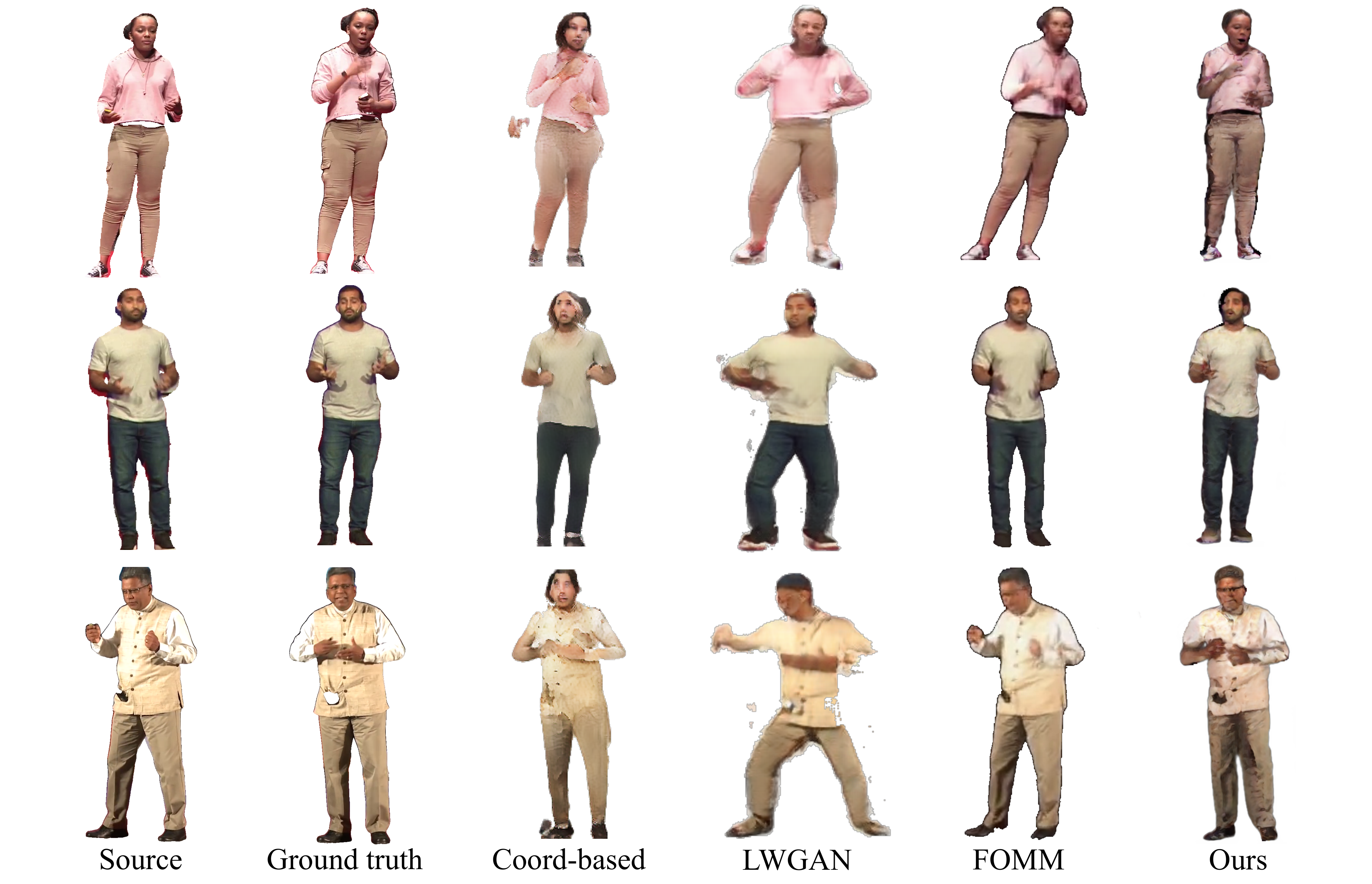}
    \caption{Comparison against state-of-the-art one-shot methods on three test sequences of the TEDXPeople dataset.}
    \label{fig:1shot_baselines}
\end{figure*}

We present comparison against state-of-the-art one-shot methods on three test sequences of the TEDXPeople dataset in Figure \ref{fig:1shot_baselines} and to PIFu and PIFuHD methods on two people from PeopleSnapshot dataset in Figure \ref{fig:1shot_baselines}. 

\begin{figure*}
    \centering
    \includegraphics[width=\textwidth]{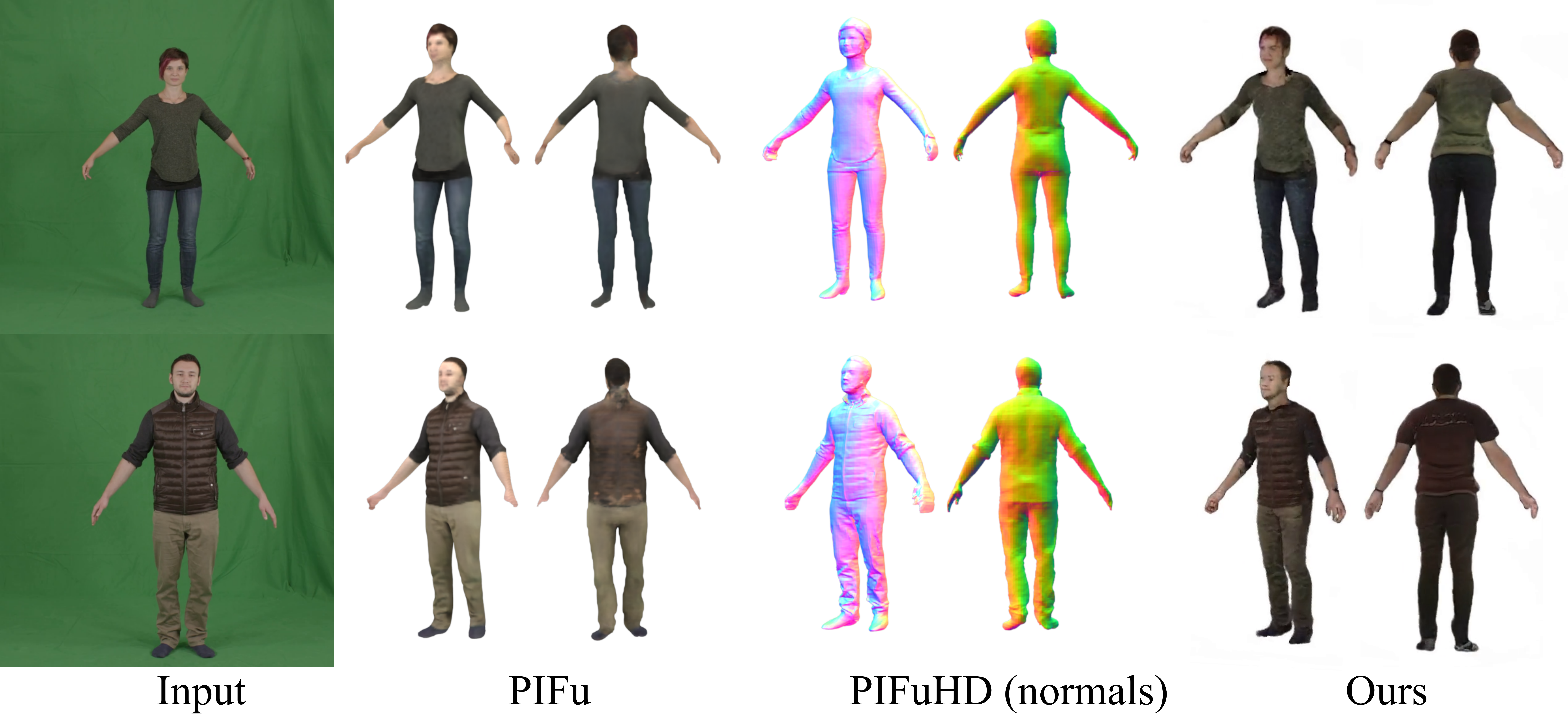}
    \caption{Comparison of our model with PIFu and PIFuHD in a one-shot mode. Normals are used for the visualization of PIFuHD since their system does not infer texture. Digital zoom is recommended. Note that unlike PIFu and PIFuHD, our model produces rigged avatar and does not require 3D models with clothing/hair during training. 
    }
    \label{fig:pifu_pifuhd_comparison}
\end{figure*}

\section*{F. Inference details}
\subsection*{Encoders.}
\begin{figure*}
    \centering
    \includegraphics[width=\textwidth]{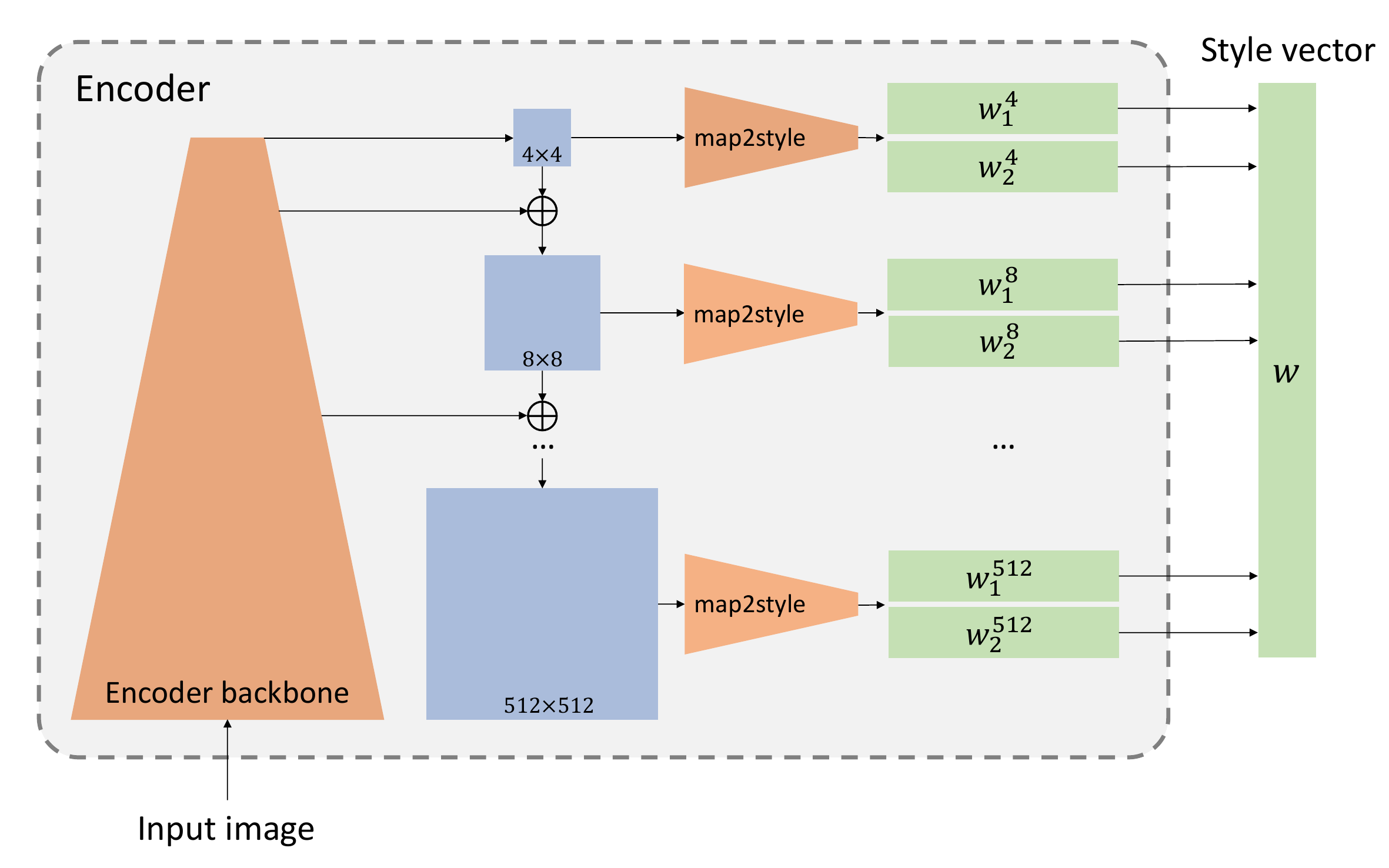}
    \caption{The encoder architecture inspired by~\cite{richardson2020encoding}. The input image is passed through the encoder convolutional backbone to extract intermediate feature maps of different resolutions ($4 \times 4, 8 \times 8, \ldots, 512 \times 512$). Each feature map except the last one is summed up with the upscaled map from the previous level and passed through a small \textit{map2style} network to predict style vectors at the corresponding generator resolution. }
    \label{fig:encoder_architecture}
\end{figure*}

The encoder architecture used in all our experiments (Fig.~\ref{fig:encoder_architecture}) is inspired by the pSp-architecture proposed in~\cite{richardson2020encoding}. As a backbone we utilize EfficientNet-B7~\cite{tan2019efficientnet} initialized with ImageNet~\cite{russakovsky2015imagenet} pretrained weigths. Intermediate feature maps are aggregated with the upscaled lower resolutions maps and then passed through a small \textit{map2style} network comprising Conv2d $\rightarrow$ ReLU $\rightarrow$ Adaptive Average Pooling $\rightarrow$ Linear layers.

We train encoders via backpropagation with ADAM optimizer~\cite{kingma2014adam}. The learning rate is fixed to $0.001$ during training.

\subsection*{Optimization.}
The optimization part of the inference consists of four stages: (i) the optimization of the latent vector $w$, (ii) the optimization of the generator parameters $h_\psi$, (iii) the optimization of the noise tensors $\mathbf N$ and (iv) the direct texture optimization. Each stage is defined by the number of iterations, the learning rate and the loss weights. Table~\ref{table:optimization_hyperparameters} summarizes hyperparameters for each stage that we found to work best.

\begin{table}[h!]
\centering
\resizebox{0.46\textwidth}{!}{
 \begin{tabular}{||l | c | c | c | c||} 
 \hline
  & Stage 1 & Stage 2 & Stage 3 & Stage 4 \\
  \hline
  \hline
  
  Number of iterations & 100 & 70 & 50 & 100 \\
  \hline
  Learning rate & 0.01 & 0.01 & 0.1 & 0.15 \\
  \hline
  \hline
  LPIPS & 1.0 & 1.0 & 1.0 & 1.0 \\
  \hline
  MSE & 0.5 & 0.5 & 0.5 & 0.5  \\
  \hline
  Encoder deviation MAE & 0.1 & 0.0 & 0.0 & 0.0  \\
  \hline
  Generator parameters deviation MAE & 0.0 & 1.0 & 0.0 & 0.0  \\
  \hline
  Texture deviation MAE & 0.0 & 0.0 & 0.0 & 0.1  \\
  \hline
  Face LPIPS & 0.5 & 1.0 & 0.1 & 2.0  \\
  \hline
  Face discriminator feature matching & 0.0 & 2.0 & 0.0 & 3.0   \\
  \hline
\end{tabular}
}
\caption{Optimization hyperparameters during inference. Stages: the optimization of latent vector $w$, the optimization of generator parameters $h_\psi$, the optimization of noise tensors $\mathbf N$ and the direct optimization of texture.}
\label{table:optimization_hyperparameters}
\end{table}

\section*{G. Ablation Study}
\subsection*{Inference ablations}
We evaluate the contributions of different losses in the inference process. With each ablations we  turn  off  one  of  the following losses: LPIPS-loss~\cite{zhang2018unreasonable} between recovered image and input image, Mean Squared Error (MSE) between recovered image and input image, Mean Absolute Error (MAE) on latent variables $w$ deviation from the initialization predicted by the encoder (MAE-encoder), MAE on generator parameters $h_\psi$ deviation from the initial ones (MAE-generator), MAE on texture deviation from the texture values (MAE-texture), LPIPS-loss~\cite{zhang2018unreasonable} on face regions of recovered and input images (LPIPS-face), feature matching loss based on trained face discriminator~\cite{Pan20} (fm-face). We compare the ablations visually and quantitatively using four metrics: Inception Score (IS), Frechet Inception Distance (FID), Learned Perceptual Image Patch Similarity (LPIPS), structural similarity index measure (SSIM).

\begin{table}[h!]
\centering
\resizebox{0.46\textwidth}{!}{
 \begin{tabular}{||l c c c c||} 
 \hline
  & IS$\uparrow$ & FID$\downarrow$ & LPIPS$\downarrow$ & SSIM$\uparrow$ \\
 \hline\hline
 \multicolumn{5}{||c||}{G-encoder} \\
  \hline
  Ours (full) & \textbf{1.8324} & 244.8 & \textbf{0.0777} & 0.9131 \\
  \hline
  No MAE-texture & 1.7768 & \textbf{223.3} & 0.0790 & \textbf{0.9134} \\
  \hline
  No fm-face & 1.8258 & 225.8 & 0.0780 & 0.9133 \\
  \hline
  No LPIPS-face & 1.8311 & 225.7 & 0.0794 & 0.9130 \\
  \hline
  No MAE-generator & 2.006 & 270.8 & 0.0803 & 0.9122 \\
  \hline
  No MAE-encoder & 1.8457 & 239.9 & 0.0797 & 0.9127 \\
  \hline
  No MSE & 1.8314 & 278.1 & 0.0809 & 0.9061 \\
  \hline

  \multicolumn{5}{||c||}{A-encoder} \\
  \hline
  Ours (full) & \textbf{1.8077} & 231.3 & \textbf{0.0785} & \textbf{0.9135} \\
  \hline
  No MAE-texture & 1.7414 & 229.4 & 0.0794 & 0.9131 \\
  \hline
  No fm-face & 1.7758 & \textbf{226.0} & 0.0799 & 0.9125\\
  \hline
  No LPIPS-face & 1.7177 & 234.4 & 0.0794 & 0.9134 \\
  \hline
  No MAE-generator & 1.7626 & 252.4 & 0.0812 & 0.9129 \\
  \hline
  No MAE-encoder & 1.6809 & 252.5 & 0.0820 & 0.9129 \\
  \hline
  No MSE & 1.7176 & 338.1 & 0.0833 & 0.9096 \\
  \hline
\end{tabular}
}
\caption{Ablation study for inference module of neural textures (one-shot). We use two sequences of the People Snapshot dataset~\cite{alldieck2018video}, taking the first frame as a source image and evaluating on the remaining frames of the sequences. See text for discussion. }
\label{table:inference_ablations_1shot}
\end{table}

\begin{figure*}
    \centering
    \includegraphics[width=\textwidth]{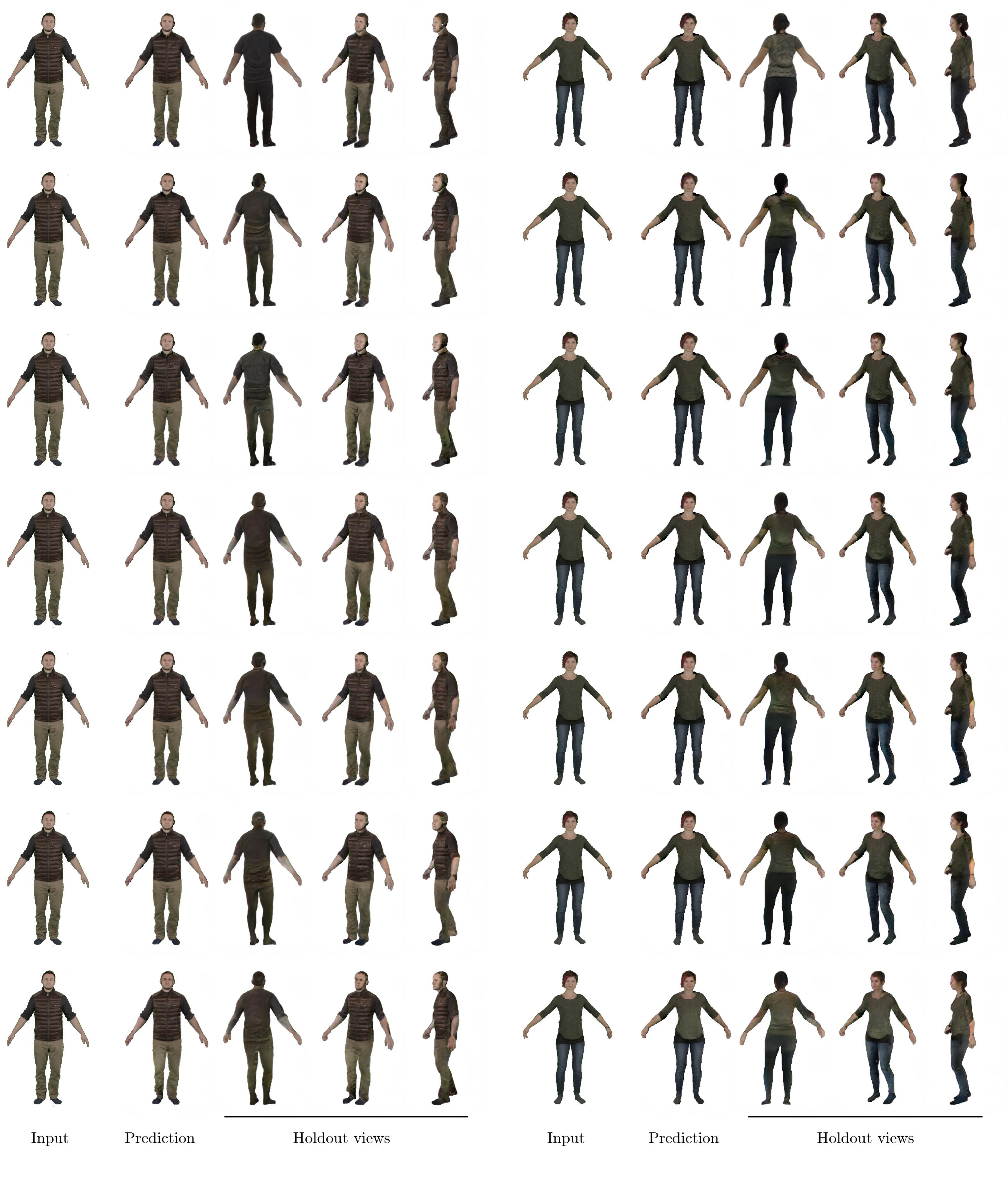}
    \caption{Ablation study for inference module of neural textures. Upper row: full model with all losses, in the following lines we are eliminating losses one-by-one following the order from Table~\ref{table:inference_ablations_1shot}.}
    \label{fig:redressing}
\end{figure*}

\section*{H. Additional experiment}
\subsection*{Redressing} 
Our approach allows the avatars to try on different clothes. This can be done with a hybrid model combining the head texture from one avatar and the non-head texture parts from the other avatar. The results of such a redressing for three randomly selected Azure people are shown in Figure~\ref{fig:redressing}.

\begin{figure*}
    \centering
    \includegraphics[width=\textwidth]{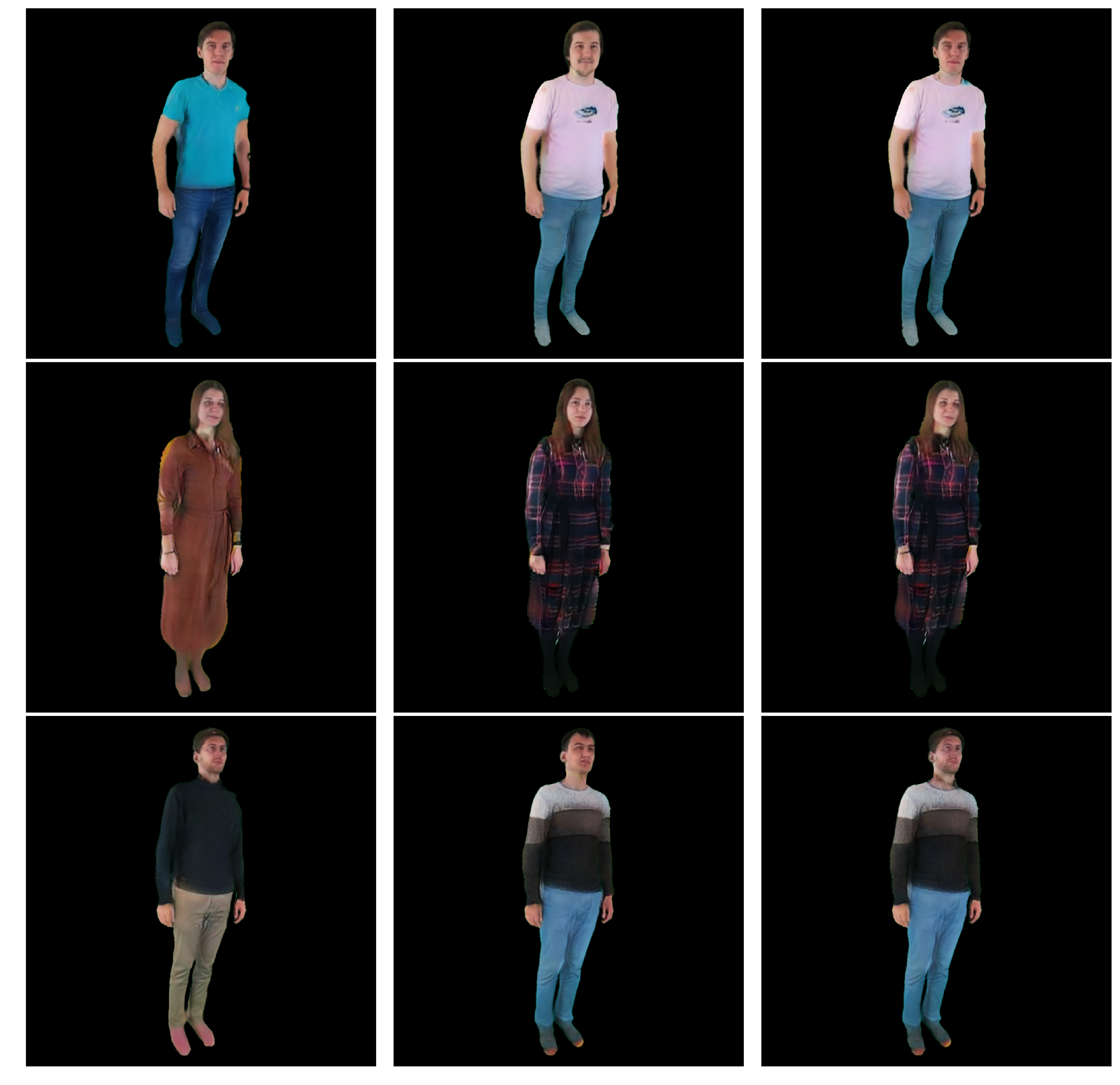}
    \caption{\textbf{Neural redressing}. Given two avatars (left and middle), we create a hybrid (‘’redressed'') avatar by taking the head neural texture from the first avatar and non-head neural texture from the second avatar. }
    \label{fig:redressing}
\end{figure*}

\end{document}